# Assistive Diagnostic Tool for Brain Tumor Detection using Computer Vision

Sahithi Ankireddy
James B. Conant High School
Hoffman Estates, USA
sahithia14@gmail.com

*Abstract—* Today, over 700,000 people are living with brain tumors in the United States. Brain tumors can spread very quickly to other parts of the brain and the spinal cord unless necessary preventive action is taken. Thus, the survival rate for this disease is less than 40% for both men and women. A conclusive and early diagnosis of a brain tumor could be the difference between life and death for some. However, brain tumor detection and segmentation are tedious and time-consuming processes as it can only be done by radiologists and clinical experts. The use of computer vision techniques, such as Mask R Convolutional Neural Network (Mask R CNN), to detect and segment brain tumors can mitigate the possibility of human error while increasing prediction accuracy rates. The goal of this project is to create an assistive diagnostics tool for brain tumor detection and segmentation. Transfer learning was used with the Mask R CNN, and necessary parameters were accordingly altered, as a starting point. The model was trained with 20 epochs and later tested. The prediction segmentation matched 90% with the ground truth. This suggests that the model was able to perform at a high level. Once the model was finalized, the application running on Flask was created. The application will serve as a tool for medical professionals. It allows doctors to upload patient brain tumor MRI images in order to receive immediate results on the diagnosis and segmentation for each patient.

## I. Introduction

Brain tumors affect all men, women and children regardless of race or ethnicity [1]. Unless proper action is taken, brain tumors can spread rapidly to other parts of the brain. Therefore, the survival rate is less than 40% for both men and women. A conclusive and early diagnosis of a brain tumor could be the difference between life and death for some. Yet, brain tumor detection and segmentation are painstaking tasks because they can only be done by radiologists and clinical experts. The accuracy of a diagnosis is based on experience solely. It is imperative to find a solution as the number of brain tumor patients are expected to rise in the near future. The use of computer vision techniques, such as Mask R Convolutional Neural Network, to detect and segment brain tumors can attenuate the possibility of human error while increasing prediction accuracy rates. The expected outcome of the use of these data analytic techniques is a higher prediction and segmentation accuracy. Furthermore, the best version of the model will be used to create a brain tumor detection and segmentation application. Users can upload a brain MRI scan, and then the application will read the image to determine a reliable brain tumor diagnosis with segmentation. This tool can possibly be a great contribution to the medical field as it can be used by doctors to assist them in diagnosing patients.

## II. Background Information

With the rise in big data, machine learning (ML) has become a key method and technique for solving problems in various areas such as computational biology and finance, computer vision, aerospace and manufacturing. ML is a general data analysis technique that uses computational models and methods to "learn" information directly from data without using a preset formula or rule-based programming. ML algorithms teach and train computers to recognize patterns and correlations between data [2]. The algorithms are able to improve their performance as the number of data points increase. A subfield of machine learning and artificial intelligence includes computer vision. This involves developing algorithms to help computers understand the content of digital images or videos [3]. Convolutional neural networks (CNNs) are special algorithms designed to take in an image and assign various weights in order to differentiate between images. CNNs consist of neurons with learnable weights and biases. Unlike a regular neural net, a CNN has its neurons arranged into 3 dimensions: width, height and depth. The neurons will only be connected to a small region of the layer before it. The last output layer "will reduce the full image into a single vector of class scores, arranged along the depth dimension "[4].

Image Segmentation creates a bounding box/mask around each specific object in an image. This technique allows for a much more granular understanding of the items in the image. There are 2 types of image segmentation: Semantic and Instance. Semantic segmentation will classify all the objects of a class as a single instance. Instance segmentation classifies objects of the same class as different instances (refer to Figure 1).

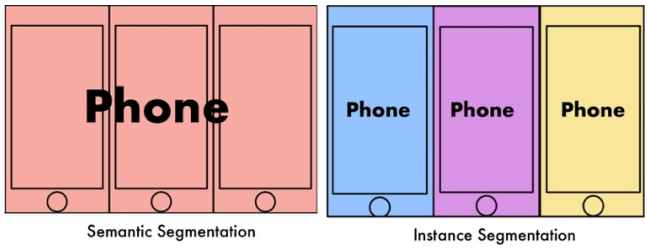

Fig. 1. Representation of Semantic and Instance Segmentation

A convolutional neural network alone will classify an image into different types. However, with segmentation, one can determine the exact location of the object in the picture. A Mask R-CNN can be used for segmentation. This model's architecture is overlaid and built on top of the Faster R-CNN. A Faster RCNN uses a CNN to extract attribute maps from the images. The maps act as inputs for the next layer and are then passed through a Region Proposal Network (RPN), which determines the candidate's bounding boxes. The RPN predicts if an object is in a certain region or not. Next, a Region of Interest (RoI) pooling layer is applied on the regions, obtained from the RPN, in order to convert all the areas to the same shape. Finally, the regions are passed through a network in order to predict the class and specific location/bounding boxes. In addition to all of this, a Mask R-CNN also creates a segmentation mask, therefore differing from the Faster R-CNN. A mask branch is added to the model's architecture. This returns a mask for every region that contains an object. Furthermore, as the Mask R-CNN is a very complex algorithm, training is time-consuming. Thus, in order to reduce training time for the Mask R-CNN, the region of interest is pre-computed. To do this, an Intersection over the Union (IoU) score is determined (refer to Figure 2).

$$IoU = \frac{Area\ of\ Overlap}{Area\ of\ Union}$$

Fig. 2. Representation of IoU

If and only when the IoU is greater than 0.5, the region is considered and passed through the model. If not, then the region is ignored. This can greatly reduce computing time, as the number of regions to evaluate is reduced [5]. Transfer learning is a popular method that can decrease computing time and resources while training models. Transfer learning is where a pre-trained model for one task is re-purposed for another task. The knowledge of an existing model is transferred and changed to meet the goals of another specific problem. Transfer learning may improve a model because it has a higher initial skill, rate of improvement, and converged skill [6]. Refer to figure 3 below.

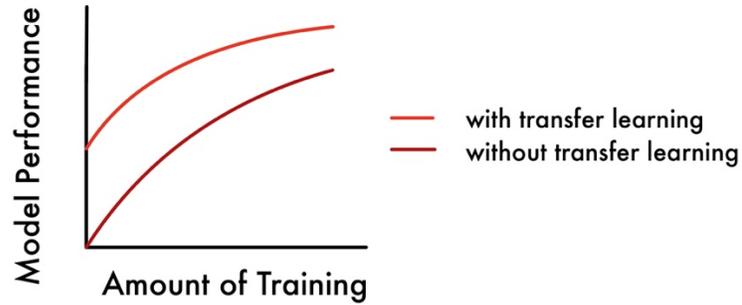

Fig. 3. Transfer and No-Transfer Learning performance chart

III. METHODS AND MATERIALS

*A. Materials*

Data pre-processing, training, and testing was all completed through the Python programming language via an interactive python Jupyter notebook. Furthermore, Anaconda, a package manager system, had to be downloaded to set up the environment. Due to the complexity of the model, training on a local computer lacked sufficient computing power. Therefore, the model was trained by a 16 vCPUs and 60 GB memory Linux Server running Ubuntu, hosted by Google Cloud Services. A publicly available supervised data set provided was used to train the model. This data set contains a total of 310 brain MRI scans. 155 scans have a brain tumor 155 in each class (yes and no). See figure 4 for a sample image.

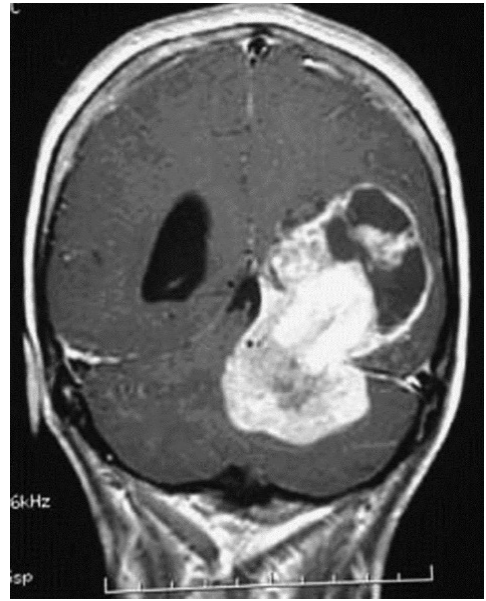

Fig. 4. Sample Brain MRI Scan

For data pre-processing and visualization, numpy and matplotlib were used. The model was engineered through Keras, a neural network framework.

## B. Model Generation and Training

It's very difficult to train a segmented CNN model from start up with randomly initialized weights. Thus, transfer learning was employed to obtain pre-learned features from the Mask R CNN. The Mask R CNN configuration is extended, and parameters are changed to match the brain tumor detection problem. Next, the Mask R CNN dataset class is extended to execute certain methods such as loading the brain scan dataset, mask and etc. After the primary steps were completed, the model was prepared to be trained. The model directory, weights and configuration were put together. Furthermore, the testing, training and validation sets were determined. The model was accordingly trained with 20 epochs. In order for the most accurate testing, the model was recreated using inference mode. This picks up the last iteration of the model trained earlier, the most updated version, for prediction. Finally, the predictions were displayed including the segmentation on the image. Refer to Figure 5 below for a flow chart of the entire process.

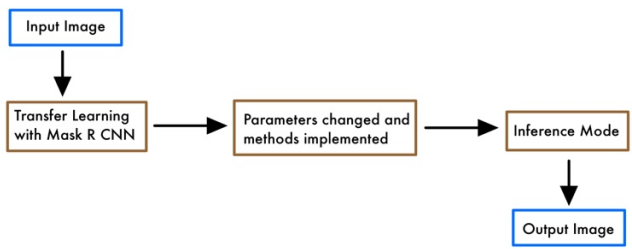

Fig. 5. Implementation Flow Chart

## C. Web Application

Once the model was evaluated and finalized, the assistive diagnostic application, running on Flask, was created. The application will serve as a tool for medical professionals. It allows doctors to upload patient brain tumor MRI images in order to obtain immediate results on the diagnosis and segmentation for each patient. The application was created using Python Flask REST API. First, the model was exported and saved, and the user interface was developed using Bootstrap. Logins and file upload features were added. After the user uploads a brain MRI scan, it's put into a location defined in the program. Once saved to the correct location, the image is passed to the Mask R CNN model for brain tumor diagnosis and segmentation. Additionally, the patient ID column was initially removed as it's not needed for the diagnosis through the model. At the end, it was later appended for displaying the results. Figure 6 and 7 delineate the input and output screen of the application.

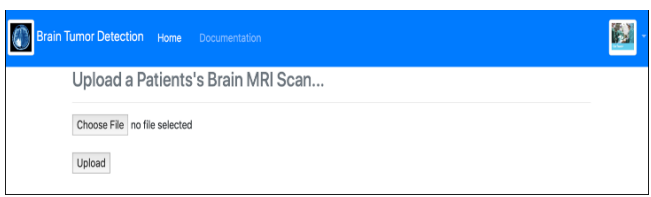

Fig. 6. Depiction of Input Screen

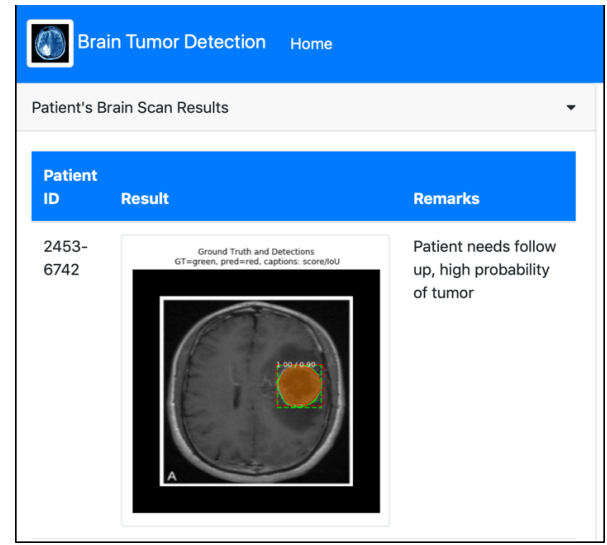

Fig. 7. Depiction of Output Screen

## IV. RESULTS AND DISCUSSION

In Figure 8, three original images and their corresponding predicted segmentation is shown. As one can see, the model correctly segmented the location of the tumor across all three images. The calculated IoU score was consistently over 0.5 for the example images in figure 8 as well as the rest. Thus, the ground truth and predictions are similar to each other, signifying that the model was able to perform to a high level.

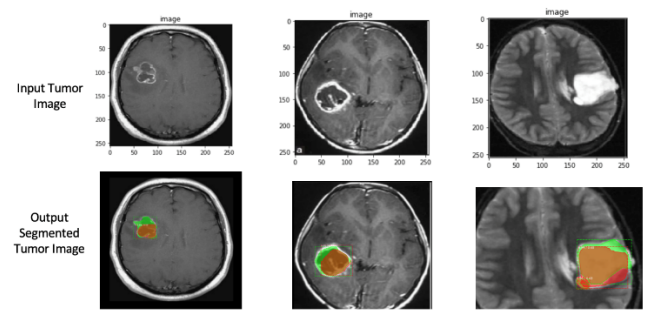

Fig. 8. Example input and output images

Shown in figure 9 is the precision-recall curve for the model. Precision is the ratio of correctly predicted positive values to the total predicted positive values. It measures how skilled a model is at predicting the positive class. Recall is the number of true positives divided by the sum of true positives and false negatives. The graph in figure 9, to a moderate extent, bows towards (1, 1) indicating a decent trained model.

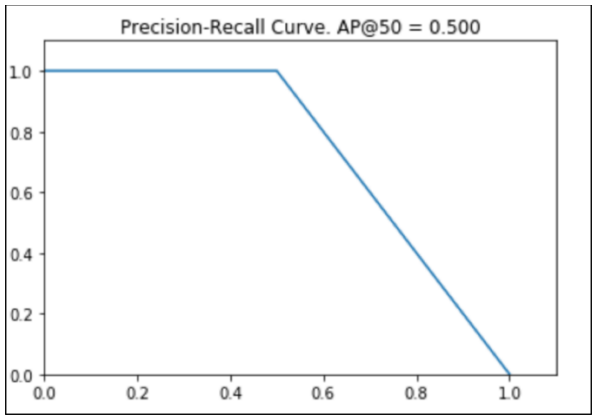

Fig. 9. Precision Recall Curve

As seen in figure 10, the loss generally decreases as the epochs increase. This illustrates that the model can backpropagate and change the weights well.

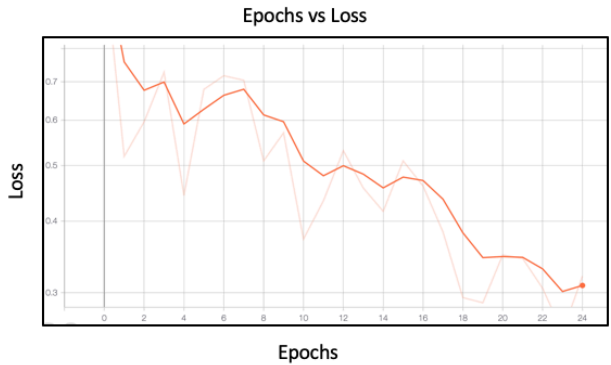

Fig. 10. Epochs vs. Loss Line Graph

Moreover, the mean average precision (mAP) was calculated. mAP computes the average precision value for recall value over 0 to 1. The model had an average mAP score of 0.60. The average intersection over union, which looks at area of overlap over area of the union, was 0.90. Both values are greater than 0.5, highlighting the model's high performance. Overall, the model's ground truth and predictions matched with the IoU score of 0.90, well over the threshold of 0.5. This demonstrates that 90% percent of the ground truth for an image matched the prediction generated by the model. The precision recall curve bowed to (1,1), and the loss function decreased as epochs increased, stipulating no sign of overfitting. Additionally, the mAP score was over 0.5. All these metrics substantiate the high skill of the model.

## V. FUTURE STEPS

Future work includes using Augmented Reality (AR) to allow physicians to incorporate real time data visualization during surgical procedures. AR provides the ability to see inside the patient without deep incisions, greatly reducing the risk during surgery [7]. Additionally, doctors will not need to depend on screens as much during surgery. Their eyes are on the patient the entire time, decreasing more risk as well. To do this, DICOM 2D images will be converted to 3D, and then the 3D model will be rendered in Unity to create the AR experience. Registration will occur to ensure the tumor will always be at the correct coordinate . Finally, AR glasses such as the Microsoft HoloLens will serve as a medium for the AR experience to be used before and during surgery. Overall, real time 3D visualization through AR along with the current brain tumor application will allow for a highly useful tool that can detect and segment in real time during surgery

## VI. ACKNOWLEDGMENT

I want to express my gratitude to Iordanis Fostiropoulos, a PhD student at USC and my mentor who guided me through my research. I also want to thank Mr. Rishu Garg for his comments, constructive criticism and advice. Lastly, I appreciate Mr. Adi Kadimetla's help in setting up Google Cloud.

## VII. REFERENCES

[1] "Quick Brain Tumor Facts." *National Brain Tumor Society*, braintumor.org/brain-tumor-information/brain-tumor-facts/.

[2] What Is Machine Learning? | How It Works, Techniques & Applications. (n.d.).Retrieved January 12, 2019, from https://www.mathworks.com discovery/machine-learning.html#how-it-work

[3] Brownlee, J. (2019, March 19). A Gentle Introduction to Computer Vision. Retrieved January 12, 2019, from https://machinelearnin gmastery.com/what-is-computer-vision/

[4] "Convolutional Neural Networks." *CS231n Convolutional Neural Networks for Visual Recognition*, cs231n.github.io/convolutional-networks/.

[5] Sharma, Pulkit. "Step-by-Step Implementation of Mask R-CNN for Image Segmentation." *Analytics Vidhya*, 22 July 2019, www.analyticsvidhya.com/blog/2019/07/computer-vision-implementing-mask-r-cnn-image-segmentation/.

[6] Brownlee, Jason. "A Gentle Introduction to Transfer Learning for Deep Learning." *Machine Learning Mastery*, 16 Sept. 2019, machinelearningmastery.com/transfer-learning-for-deep-learning/.

[7] "What Is Augmented Reality?" *The Franklin Institute*, 18 Dec. 2019, www.fi.edu/what-is-augmented-reality.